\documentclass[10pt, a4paper]{article}

\usepackage[final]{lrec-coling2024} 
\usepackage[linesnumbered,ruled,vlined]{algorithm2e}
\usepackage{algorithmic}
\usepackage{tabularx}
\usepackage{inconsolata}
\usepackage{makecell}
\usepackage{soul}
\usepackage{lineno}
\usepackage{enumitem}
\usepackage[most]{tcolorbox}
\newtcolorbox{mybox}[2][]{
  fonttitle=\bfseries,
  title=#2,
  #1
}
\usepackage[linesnumbered, ruled]{algorithm2e}
\setenumerate[1]{itemsep=1pt,partopsep=0pt,parsep=\parskip,topsep=5pt}
\setitemize[1]{itemsep=0pt,partopsep=0pt,parsep=\parskip,topsep=5pt}
\setdescription{itemsep=0pt,partopsep=0pt,parsep=\parskip,topsep=5pt}
\usepackage{multirow}
\usepackage{amsmath}
\usepackage{booktabs}
\usepackage{natbib}
\usepackage{multibib}
\title{Self-Improvement Programming for Temporal Knowledge Graph Question Answering}

\name{Zhuo Chen$^{1,2}$, Zhao Zhang$^{1,2}$, Zixuan Li$^{*,1,2}$ \thanks{*Corresponding author}, Fei Wang$^{\ast,1,2}$, Yutao Zeng, \\{\bf \large Xiaolong Jin$^{1,2}$ and Yongjun Xu$^{1,2}$}
} 
\address{$^{1}$Institute of Computing Technology, Chinese Academy of Sciences, Beijing, China \\ 
$^{2}$University of Chinese Academy of Sciences, Beijing, China \\ 
         \{ chenzhuo23s, zhangzhao2021, lizixuan, wangfei, jinxiaolong, xyj \}@ict.ac.cn
\\
         {yutao.zeng}@outlook.com}

\abstract{
Temporal Knowledge Graph Question Answering (TKGQA) aims to answer questions with temporal intent over Temporal Knowledge Graphs (TKGs). The core challenge of this task lies in understanding the complex semantic information regarding multiple types of time constraints (e.g., before, first) in questions. Existing end-to-end methods implicitly model the time constraints by learning time-aware embeddings of questions and candidate answers, which is far from understanding the question comprehensively. 
Motivated by semantic-parsing-based approaches that explicitly model constraints in questions by generating logical forms with symbolic operators, we design fundamental temporal operators for time constraints and introduce a novel self-improvement \textbf{Prog}ramming method for \textbf{T}KG\textbf{QA} (\textbf{Prog-TQA}). 
Specifically, Prog-TQA leverages the in-context learning ability of Large Language Models (LLMs) to understand the combinatory time constraints in the questions and generate corresponding program drafts with a few examples given. Then, it aligns these drafts to TKGs with the linking module and subsequently executes them to generate the answers. To enhance the ability to understand questions, Prog-TQA is further equipped with a self-improvement strategy to effectively bootstrap LLMs using high-quality self-generated drafts. 
Extensive experiments demonstrate the superiority of the proposed Prog-TQA on MultiTQ and CronQuestions datasets, especially in the Hits@1 metric.
 \\ \newline  }

\begin{document}

\maketitleabstract

\section{Introduction}\label{sec:intro}
Temporal Knowledge Graphs (TKGs) store real-world facts along with a timestamp or time interval   (\emph{e.g.}, \textit{(China, Host a visit, Vietnam, 2015-12-27)}). Temporal Knowledge Graph Question Answering (TKGQA) aims to answer natural language questions with temporal intent (i.e., temporal questions) over TKGs. The task is particularly important in various applications, such as historical research, financial analysis, and understanding trends over time, where the temporal intent is crucial for accurate and meaningful answers~\cite{ludemann2020knowledge, xu2023artificial}.

Compared to common questions, temporal questions engage multiple time constraints and thus have more complex semantic information. The core challenge of the TKGQA task is how to understand semantic information comprehensively by diving deep into the combinatory time constraints in the question. 
For example, the temporal question ``\textit{Which country was last visited by Obama before he visited the Foreign Minister of Cape Verde?}” involves dual time constraints ``\textit{before}'' and ``\textit{last}''. By understanding the time constraints, the question can be answered by the following reasoning steps: 1) Localize the time point in terms of the fact ``\textit{(Obama visit, Foreign Minister of Cape Verde, 2013-03-28)}''; 2) Query the facts about \textit{Obama's visit}; 3) Filter the facts preceding the above time point according to the ``\textit{before}’' constraint; 4) Sort the filtered facts and get the country in the latest fact according to the ``\textit{last}'’ constraint.

Existing TKGQA methods~\cite{saxena2021question,mavromatis2022tempoqr} implicitly model the time constraints by learning time-aware embeddings of both questions and candidate answers. Then, they predict answers by calculating the similarity between these embeddings. Nevertheless, such methods struggle to model and utilize time constraints precisely, further making them unable to fully understand the complex semantic information of combinatory constraints in temporal questions.

Different from these methods, semantic-parsing-based approaches~\cite{chen2021retrack, gu2022arcaneqa, shu2022tiara} in Knowledge Graph Question Answering (KGQA) task convert natural language questions to logical forms (e.g., SPARQL queries), and subsequently execute them on Knowledge Graphs (KGs) to retrieve answers. Such approaches parse the semantics of questions into the composition of symbolic operators. These operators are capable of handling numerical constraints, comparisons, and ordering, thus offering a feasible solution to answer temporal questions.
\begin{figure}
    \centering
    \includegraphics[width=1\linewidth]{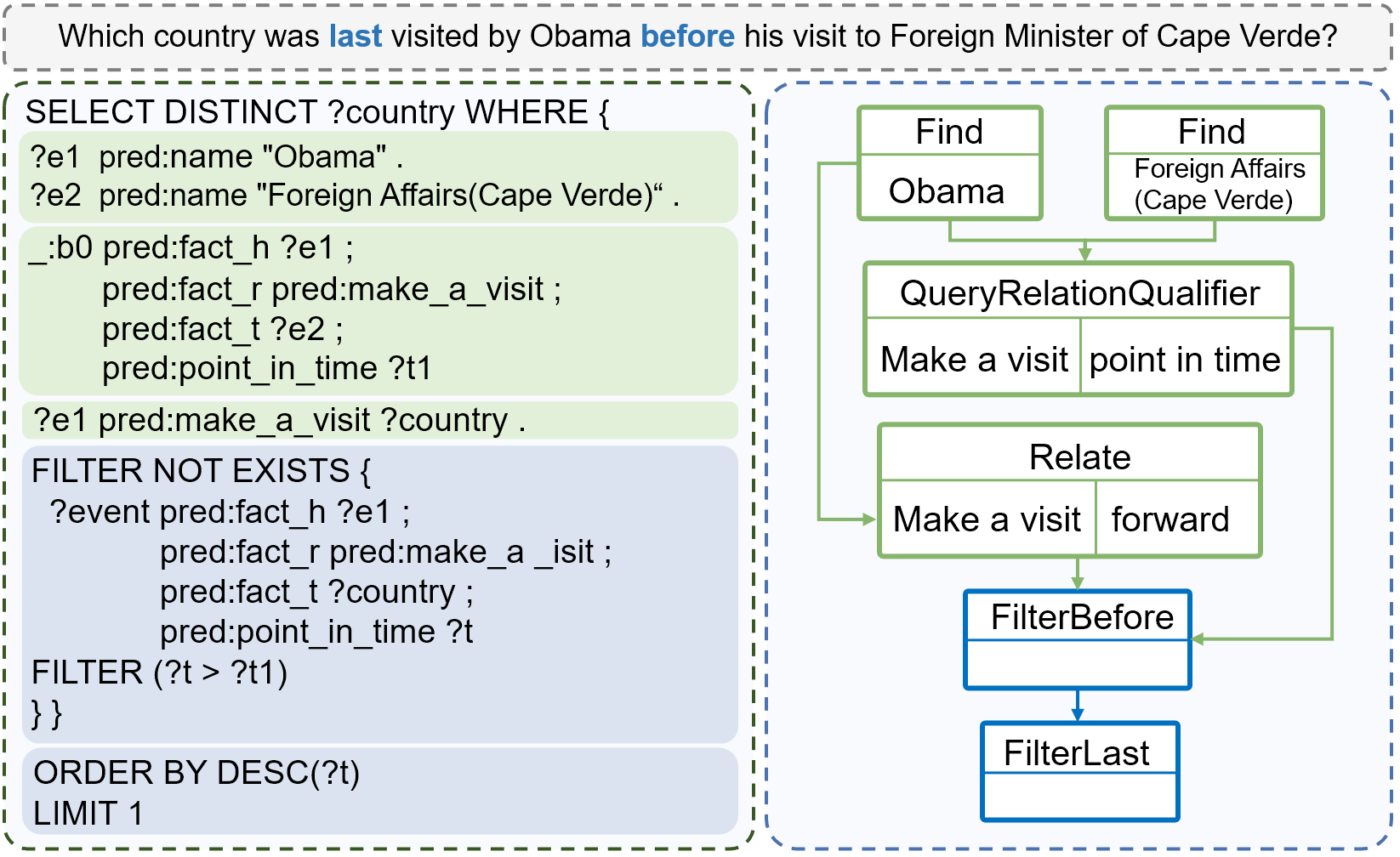}
    \caption{SPARQL query with complex clauses for temporal operations (left) and extended KoPL with neat temporal operators (right).} 
    \label{fig:kopl-example}
    \vspace{-1em}
\end{figure}
However, applying existing semantic-parsing-based methods directly to temporal questions encounters two challenges:  1) Lack of brief temporal operators: Common query languages (e.g., SPARQL) lack specifically designed temporal operators. As the left part of Figure~\ref{fig:kopl-example} shows, the SPARQL query represents temporal constraints through ordinary numerical operators in the form of multiple query clauses (depicted in blue), which are not concise enough for automated generation; 2) Lack of logical form annotations: Traditional semantic-parsing-based methods heavily rely on extensive annotations for training. Without sufficient annotated logical forms for temporal questions, it's hard to equip models with abilities to parse complex questions into query programs with temporal operators.

To overcome the first challenge, we systematically analyze time constraints and design corresponding temporal operators. Specifically, we choose the Knowledge-oriented Programming Language (KoPL)~\cite{cao2022kqa}, which offers extensibility and adaptability, extending its function library by implementing the designed temporal operators. As illustrated in Figure \ref{fig:kopl-example}, performing a single temporal operation in the extended KoPL requires only one temporal operator, whereas achieving the same in SPARQL necessitates multiple clauses.

Recently, LLMs (e.g., ChatGPT) have shown impressive semantic understanding and robust few-shot generalization capabilities on a variety of NLP tasks~\cite{brown2020language}. Thus, we tackle the second challenge by utilizing the In-Context Learning (ICL) ability of LLMs and propose a two-stage framework called Prog-TQA. Prog-TQA can generate programs incorporating external TKG schemas flexibly. In the first stage, Prog-TQA leverages the ICL ability of the LLM to parse questions and generate primarily program drafts according to a few question-program examples. In the second stage, Prog-TQA aligns the mentions in drafts to specific TKG using a similarity-based linking module.
To further improve the LLM's ability to understand complex temporal questions, we integrate Prog-TQA with a self-improvement strategy, enabling the LLM to refine itself utilizing self-generated programs. This process uses gold answers as weak supervision to assess programs. It progressively bootstraps LLM's ability to understand questions by iteratively learning from the correct programs and correcting the challenging questions.

In summary, our contributions are as follows:
\begin{itemize}
  \item We systematically analyze time constraints in TKGQA and design the corresponding temporal operators to extend KoPL to handle temporal questions. 
  \item Based on the designed operators, we propose a two-stage framework for the TKGQA task, namely Prog-TQA, which leverages the ICL ability to perform few-shot program generation. Besides, we incorporate an effective self-improvement strategy in Prog-TQA to enhance LLM's comprehension of complex questions.
  \item Experiments demonstrate the superiority of Prog-TQA. It achieves up to 50.4\% improvement overall at Hits@1 on MultiTQ and 3.5\% for complex questions at Hits@1 on CronQuestions.
\end{itemize}
\begin{figure*}[ht]
    \centering
    \includegraphics[width=1\linewidth]{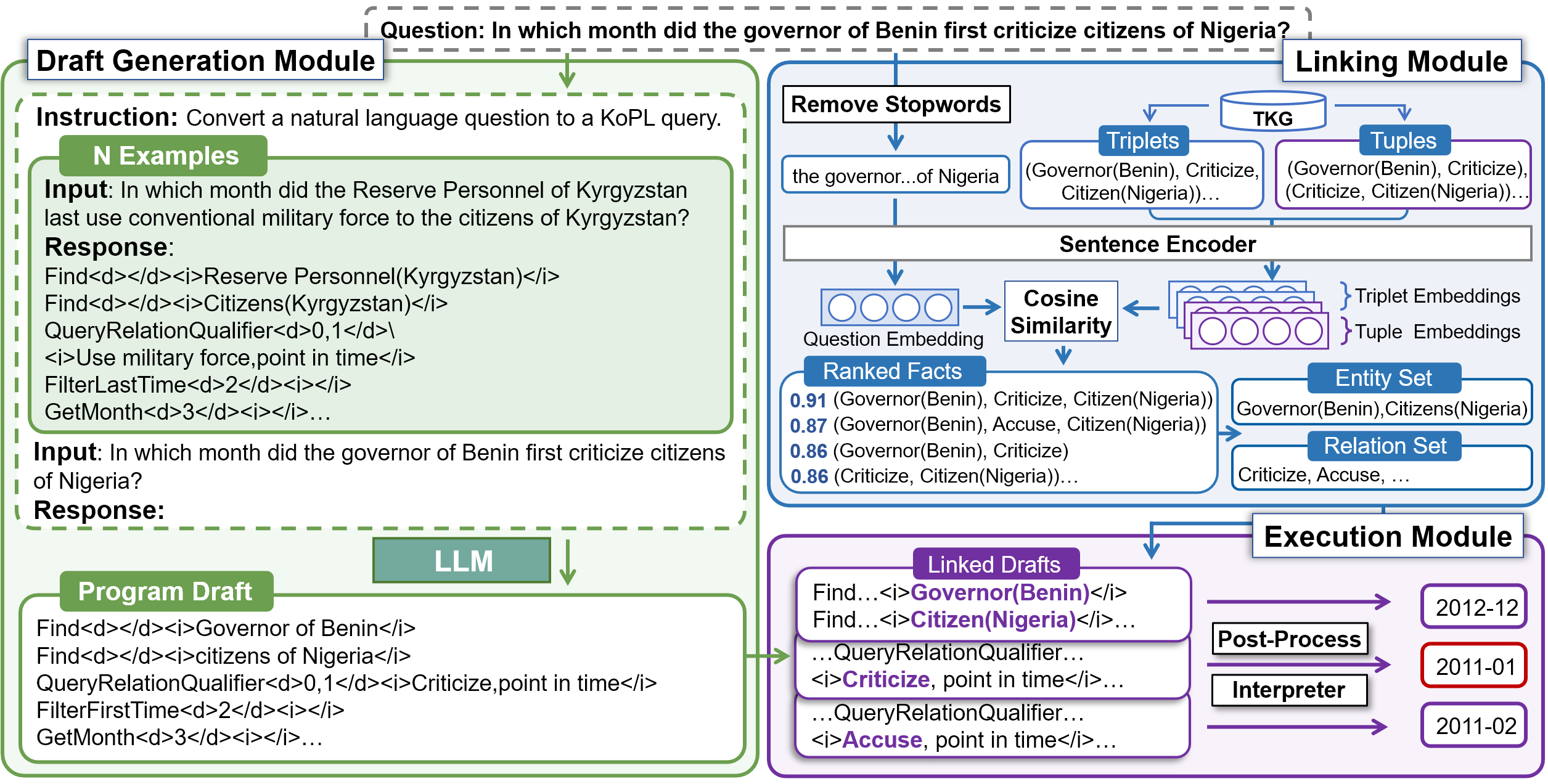}
    \caption{The pipeline of proposed Prog-TQA. Prog-TQA firstly leverages the ICL ability of LLMs to generate KoPL program drafts. Subsequently, it adopts linking and execution modules to complete and execute the generated drafts on TKG and finally obtains the answers.}
    \label{fig:model-0}
    \vspace{-1em}
\end{figure*}
\section{Related Work}
\textbf{TKGQA methods.} 
Existing TKGQA methods can be divided into two categories: the embedding-based methods and semantic-parsing-based methods. Embedding-based methods, which constitute the mainstream, typically learn time-aware embeddings for questions and candidate answers and predict answers by embedding calculations.  CronKGQA~\cite{saxena2021question} obtains entity and time embeddings from pre-trained TKG models. Building upon this, TempoQR~\cite{mavromatis2022tempoqr} retrieves the time scopes of annotated entities and integrates them into the question representation. MultiQA~\cite{chen2023multi} extracts time information and introduces a multi-granularity time aggregation module to enhance question embeddings.

The second category of approaches, which remains relatively unexplored, revolves around semantic parsing. These approaches aim to identify time constraints within questions and subsequently decompose the questions or generate corresponding query structures. TEQUILA~\cite{jia2018tequila}  decomposes each question into non-temporal and temporal sub-questions, addressing them separately. SF-TQA~\cite{ding2022semantic} organizes the time constraints into predefined structures and subsequently generates and executes them for answer retrieval. 

Different from the embedding-based methods that implicitly model time constraints, Prog-TQA explores parsing the questions into explicit query steps, and distinguishes itself from other semantic-parsing-based methods by the requirement of few annotations.

\textbf{Semantic-parsing-based KGQA methods.} Semantic-parsing-based approaches in KGQA convert the questions into logical forms and then execute them on the KGs to obtain answers. Traditional methods~\cite{chen2021retrack, gu2022arcaneqa, shu2022tiara} leverage the generative capability of Language Models (LMs), which fine-tune LMs and use them as semantic parsers. 
To eliminate the need for annotations, some recent works take advantage of the advanced LLMs. KB-BINDER~\cite{li2023few} firstly leverages the ICL ability of LLMs to generate logical forms with few annotations. McL-KBQA~\cite{tan2023make} employs ICL with multiple choices while Nie et al.~\cite{nie2023code} introduces a code-style ICL method to reduce format errors.
The proposed Prog-TQA also falls into this category. It explores the parsing of temporal questions based on designed temporal operators.

\textbf{Self-improvement.} 
Huang et al.~\cite{huang2023large} introduces the self-improvement of LLMs, leveraging self-consistency in generated reasoning paths as weak supervision. Similarly, we adopt a self-improvement strategy in Prog-TQA, using gold answers as weak supervision to distinguish generated programs. Differently, the iterative process is highlighted to solve complex temporal questions gradually.
\section{Methodology}
\subsection{Overview}
Given natural language questions with temporal intent and a TKG $\mathcal{G}:=\{(s,r,o,t), s,o \in \mathcal{E}, r\in \mathcal{R}, t \in \mathcal{T}\}$, where $\mathcal{E}$, $\mathcal{R}$ and $\mathcal{T}$ represent the set of entities, relations and time separately. The task of TKGQA aims to answer these temporal questions by referencing the facts stored in the TKG. 
To achieve this, we first design a set of fundamental temporal operators that empower KoPL to execute temporal operations (Section~\ref{sec:atq}). Building upon this foundation, we propose a novel TKGQA framework named Prog-TQA (Section~\ref{sec:Prog-TQA}).

\begin{table*}
\centering
\resizebox{\textwidth}{!}{
    \begin{tabular}{c|c|c|c}
    \toprule
    \textbf{Operator Types} & \textbf{Temporal Operators}  & \textbf{Descriptions} & \textbf{Examples (Simplified)} \\
    \midrule
    \multirow{4}{*}{Fundamental} & FilterBefore & 
    Retrieve facts occurring before the given time. & 
    FilterBefore(Facts,``2010-01-12") \\ 
    & FilterAfter & 
    Retrieve facts occurring after the given time. & 
    FilterAfter(Facts,``2010-01-12") \\
    & FilterFirst & 
    Retrieve facts or times that occur first. & 
    FilterFirst([``2010-01-12",  ``2012-11-30",``2013-11-02"]) \\
    & FilterLast & 
    Retrieve facts or times that occur last. & 
    FilterLast([``2010-01-12",  ``2012-11-30",``2013-11-02"]) \\
    \midrule
    \multirow{7}{*}{Precise} & FilterRange & 
    Retrieve facts that occur within coarse-grained time bounds. & 
    FilterRange(Facts, ``2010-10") \\
    & GetYear & 
    Get year-format answers from the time or fact list. & 
    GetYear([ ``2010-01-12",  ``2012-11-30"]) \\
    & GetMonth & 
    Get month-format answers from the time or fact list. & 
    GetMonth([ ``2010-01-12",  ``2012-11-30"]) \\
    & GetDate & 
    Get date-format answers from the time or fact list. & 
    GetDate([ ``2010-01-12",  ``2012-11-30"]) \\ 
    & GetDuration & 
    Get the duration when facts hold happening state. & 
    GetDuration(Facts) \\
    & FilterByDuration & 
    Retrieve facts that hold the happening state within the given duration(s). & 
    FilterByDuration(Facts,[``2010",``2011",``2012"]) \\
    & FilterByTimePoint & 
    Retrieve facts that hold the happening state at the given time point. & 
    FilterByTimePoint(Facts,``2012") \\
    \midrule
    Dataset-specific & QueryEventQualifier & 
    Get the given qualifiers in the ``event" type entities. & 
    QueryEventQualifier(``2004 Summer Olympics",``duration") \\
    \bottomrule
    \end{tabular} }
    \caption{Descriptions and examples for the designed operators. For ease of understanding, we provide the description and a representative example for each operator, where the input is simplified. ``Facts'' denotes the set of facts in TKGs.}
    \label{table:operators}
    \vspace{-0.5em}
\end{table*}

\subsection{The Designed Temporal Operators}\label{sec:atq}
To empower KoPL with temporal operation capabilities, we generalize three categories of temporal operations according to time constraints and implement them in KoPL's function library. Table~\ref{table:operators} provides descriptions and examples of operators, more design details can be found in Appendix \ref{app:tempop}.

\textbf{Fundamental Temporal Operators.} 
We first summarize the fundamental temporal constraints into comparative constraints (i.e., before, after, simultaneous) and ordinal constraints (i.e., first, last).  For these constraints, we develop four operators to simplify the querying process for facts occurring before or after a particular time (FilterBefore/FilterAfter), as well as to query the earliest and latest facts or time (FilterFirst/FilterLast).

\textbf{Precise temporal Operators.} 
We additionally design advanced temporal operators to enhance precise information retrieval, addressing diverse operational needs arising from nuanced temporal properties involving granularity (year, month, day) and storage format (time point and time interval).
In terms of granularity, we introduce four functions to facilitate the query of different granularity formats (GetYear/GetMonth/GetDate) and to enable the use of coarse-grained time for fact retrieval (FilterRange).
Considering the diverse storage formats of time information in TKGs, we provide operators for querying facts that occur at different format times (FilterByTimePoint/FilterByDuration).

\textbf{Dataset-Specific Operators.} 
Notably, these operators can be flexibly extended according to future requirements. To empower event queries in the CronQuestions dataset, we additionally introduce a function to query qualifiers in events, as shown in the last block.
 
\subsection{The Proposed Prog-TQA Method}\label{sec:Prog-TQA}
Based on designed temporal operators, we propose a semantic-parsing-based TKGQA method called Prog-TQA. Prog-TQA is a two-stage program generation framework, which enables the LLM to flexibly generate programs involving different external TKGs. As Figure \ref{fig:model-0} shows, Prog-TQA initially employs a draft generation module to generate a preliminary draft for a given temporal question. Subsequently, the link module aligns the draft with a specific TKG, followed by the execution module that interprets and executes the program to derive the answer. Moreover, to overcome the challenge of lacking logical form annotations, Prog-TQA further incorporates a self-improvement strategy, bootstrapping the LLM with high-quality self-generated annotations in an iterative fine-tuning manner.

\subsubsection{Draft Generation Module}\label{sec:kdg}
The draft generation module leverages the ICL ability of the LLM to understand the questions and generate program drafts with few annotations. Specifically, it first samples relevant question-program examples as demonstrations. Then, it organizes the examples to construct the prompt, which serves as the input for the LLM.

In the example retrieval process, to procure examples conducive to question answering, Prog-TQA samples $N$ examples of the same category as the query question.  To start the process, we manually annotate a very small number of demonstration examples. We subdivide the pre-classified categories based on the type of answer (i.e., entity or time) and time constraints, and then provide 20 exemplary programs annotated for each category.
 
In the prompt construction process, following the general ICL method, the prompt is organized into three main components: a brief task instruction $I$, demonstration examples $D$, and the query question $Q$. To facilitate the LLM's comprehension of the provided programs, a simplified format of the KoPL programs is given. In this format, the function name is retained. The textual arguments are encapsulated within <i></i> tags, while the functional arguments that specify the dependent functions are encapsulated within <d></d> tags. Additional details regarding annotated programs and prompt examples are available in Appendix \ref{app:annot}.

In summary, the draft generation process for the query question $Q$ using instruction $I$, selected demonstration examples $D$ can be outlined as follows:
\[
{K}_d = {M}(I;D;Q)
\]where ${K}_d$ represents for the generated KoPL program drafts, and ${M}$ represents for the LLM used.
\subsubsection{Linking Module}\label{sec:link}
The drafts generated in the first stage are not guaranteed to be executable, for the entities and relations are directly extracted from questions. To facilitate execution, the linking module aligns mentions in drafts with TKGs. 

Unlike existing methods that use a simple fuzzy matching approach, the linking module utilizes a similarity-based method to refine the linking process. It can be observed that the main entities and relations in the question semantics tend to appear in the reasoning steps to query the answers, they usually appear in the same few fact quadruples. Thus, the linking module narrows the scope of linking candidates using the semantic similarity between the core semantics of the question and the fact.

As Figure \ref{fig:model-0} shows, the module begins with skeleton extraction. It standardizes both questions and facts into a uniform sentence structure consisting mainly of entities and relations. For questions, it extracts the skeleton by removing the question words, stop words, and temporal words. As for facts, it first removes the time component to obtain triplet facts. Given that the semantics of another entity can disrupt the retrieval in questions that involve only a single entity, it also splits a triplet into two entity-relation tuples, adding extra tuple facts. Each fact is joined with spaces to form a sentence.

Then, the question and fact sentences are encoded into embeddings by the sentence encoder. The cosine similarity is utilized to select the top $k$ relevant facts, thus forming the set of entities and relations.

Finally, the linking module uses fuzzy matching to link entities and relations in program drafts with the above entity set and the relation set, respectively.
Noting that certain relations in the TKG are easily confused, the module selects $r$ relations during relation linking and generates $r$ copies of the draft.
\subsubsection{Execution Module}\label{sec:exec}
Based on the linked drafts, Prog-TQA employs an execution module to process drafts into KoPL programs and execute them to retrieve answers. Since the current program drafts are solely aligned with the TKG, they may contain parameter errors or structural inaccuracies. To address this, the execution module incorporates an optional post-processing submodule aimed at minimizing such errors before executing programs.

Errors in the parameters or structure of a program can be generalized to functional dependency errors (e.g., a functional dependency argument of 1 but set to 0) and relational active-passive argument errors (e.g., mistake ``forward" for ``backward"). The post-processing submodule enumerates all possible arguments for functions that involve functional dependency and active-passive relation arguments. Then, it produces a copy of the KoPL program for each enumeration result. 

Finally, the execution module converts the KoPL program drafts into complete programs and executes them on the KoPL engine. The final answers are obtained by combining the execution results of all programs produced for the question.
\subsubsection{Self-Improvement Strategy}\label{sec:ift}
\begin{algorithm}
\caption{Self-improvement process}
\label{table:psedocode}
\SetKwInOut{Input}{Input} 
\SetKwInOut{Output}{Output} 
\small
\Input{Training set $T$, raw LLM ${M}_0$, TKG $\mathcal{G}$, convergence threshold $C$, maximum iteration rounds $Max$}
\Output{Fine-tuned LLM ${M}_{i}$}
\SetKwFunction{MyFunction}{\textbf{GenProg}} 
\SetKwFunction{MyFunc}{\textbf{CheckProg}} 
\SetKwProg{Fn}{Function}{:}{\KwRet}
Get fine-tuning data $F_0$, gold answer $GA$ from $T$\;
$Results \leftarrow $ \MyFunction{$F_{0}$, $\mathcal{G}$, $M_0$}\;
$F_{0}^t,F_{0}^f \leftarrow $ \MyFunc{$Results, GA$}\;
$i \leftarrow 0$\;
\While{$|F_{i}^t| \geq C$ and $i < Max$}{
    ${M}_{i+1} \leftarrow $ \textbf{FineTune}($F_i^t$, ${M}_i$)\;
    
    $Results \leftarrow  $ \MyFunction{$F_{i}^f$,$\mathcal{G}$,${M}_{i+1}$}\;
    $F_{i+1}^t,F_{i+1}^f \leftarrow $ \MyFunc{$Results, GA$}\;
    $i \leftarrow i+1$\;
}
\Return ${M}_{i}$\;
\Fn{\MyFunction{$data, TKG, LLM$}}{
    $Results \leftarrow \emptyset$\;
    \ForEach{$Ques \in data$}{
        $Prog, Res \leftarrow$ \textbf{Prog-TQA}($Ques, LLM, TKG$)\;
        $Results \leftarrow Results \cup (Ques;Prog;Res)$\;
    }
\KwRet{$Results$}\;
}
\Fn{\MyFunc{$results, gold\_answer$}}{
    $Correct, Incorrect \leftarrow \emptyset, \emptyset$\;
    \ForEach{$(Ques;Prog;Res) \in results$}{
        $Flag \gets False$\;
        \ForEach{$p,r \in Prog,Res$}{
            \If{$r \cap gold\_answer \neq \emptyset$}{
                $Correct  \leftarrow Correct \cup (Ques;p)$\;
                $Flag \gets True$;
                $break$\;
            }
        }
        \If{$Flag$ is $False$}{
            $Incorrect  \leftarrow Incorrect \cup Ques$\;
        }
    }
    \KwRet{$Correct, Incorrect$}\;
}
\end{algorithm}

While Prog-TQA offers a complete program generation framework, the built-in LLM learns solely from a limited set of provided examples, constraining its capacity for intricate reasoning tasks. To address this, we incorporate Prog-TQA with a self-improvement strategy, leveraging the self-generated high-quality programs to refine the LLM. The self-improvement process involves two main stages: initially, it generates fine-tuning data using gold answers as the weak supervision; then, it iteratively conducts fine-tuning sessions, enabling the LLM to continually learn from high-quality programs and rectify challenging questions. Algorithm \ref{table:psedocode} shows the pseudo-code for the self-improvement process.

In the initial data generation stage, the fine-tuning data $F_0$ and corresponding gold answers $GA$ are sampled from the training set $T$. Then, the fine-tuning data are processed through the Prog-TQA pipeline, getting programs and execution results for each question (Algorithm \ref{table:psedocode} line
1-2 and function \verb|GenProg|). Notably, the post-processing operation is activated to over-generate candidate programs for filtration. Subsequently, the gold answers for each question serve as the weak supervision to assess programs (Algorithm \ref{table:psedocode} line
3 and function \verb|CheckProg|): for every candidate program associated with a question, if the program's execution results match the answers, the question-program pair is included in the correct dataset $F^t_0$; conversely, if no program produces matching results, indicating a challenging question, it is included in the incorrect dataset $F^f_0$.

The iterative fine-tuning process starts based on the initial fine-tuning data.  At the $i$-th iteration, the LLM ${M}_i$ undergoes fine-tuning based on $F^t_i$ to assimilate knowledge from the correct programs. Subsequently, the fine-tuned LLM ${M}_{i+1}$ is deployed to address the challenging questions in $F^f_i$. Employing the same supervision strategy, new correct and incorrect datasets $F^t_{i+1}$ and $F^f_{i+1}$ are delineated for the subsequent iteration. The iterative process persists until either the maximum number of rounds $Max$ is reached or when the size of $F^t_{i}$ decreases below a predefined threshold $C$ (Algorithm \ref{table:psedocode} line
4-10).

\begin{table*}[ht]
    \centering
    \resizebox{\textwidth}{!}{
        \begin{tabular}{l | c c c c c c c c c c}
        \toprule
        \multirow{3}{*}{\textbf{Model}} & \multicolumn{5}{c}{ Hits@1 } & \multicolumn{5}{|c}{ Hits@10 } \\ 
        \cmidrule{2-11}
        & \multirow{2}{*}{ Overall } & \multicolumn{2}{|c|}{ Question Type } & \multicolumn{2}{c|}{ Answer Type } & \multirow{2}{*}{ Overall } & \multicolumn{2}{|c|}{ Question Type } & \multicolumn{2}{c}{ Answer Type } \\
        \cmidrule{3-6}
        \cmidrule{8-11}
        & &\multicolumn{1}{|c|}{Multiple} & Single & \multicolumn{1}{|c|}{Entity} & \multicolumn{1}{c|}{Time} & & \multicolumn{1}{|c|}{Multiple} & Single & \multicolumn{1}{|c|}{Entity} & \multicolumn{1}{c}{Time} \\
        \midrule
        BERT & 0.083 & 0.061 & 0.092 & 0.101 & 0.040 & 0.441 & 0.392 & 0.461 & 0.531 & 0.222 \\
        \midrule
        EmbedKGQA & 0.206 & 0.134 & 0.235 & 0.290 & 0.001 & 0.459 & 0.439 & 0.467 & 0.648 & 0.001 \\
        CronKGQA & 0.279 & 0.134 & 0.337 & 0.328 & 0.156 & 0.608 & 0.453 & 0.671 & 0.696 & 0.392 \\
        MultiQA & \underline{0.293} & \underline{0.159} & \underline{0.347} & \underline{0.349} & \underline{0.157} & \underline{0.635} & \underline{0.519} & \underline{0.682} & \underline{0.733} & \underline{0.396} \\
        \midrule
        Prog-TQA & \textbf{0.797} & \textbf{0.750} & \textbf{0.817} & \textbf{0.790} & \textbf{0.815} & \textbf{0.934} & \textbf{0.910} & \textbf{0.944} & \textbf{0.922} & \textbf{0.963} \\
        \bottomrule
        \end{tabular}
        }
    \caption{Comparison of various baselines and Prog-TQA on MultiTQ.
    }
    \label{table:multiQA}
\end{table*}
\begin{table*}[ht]
    \centering
    \resizebox{\textwidth}{!}{
        \begin{tabular}{l | c c c c c c c c c c}
        \toprule
        \multirow{3}{*}{\textbf{Model}} & \multicolumn{5}{c}{ Hits@1 } & \multicolumn{5}{|c}{ Hits@10 } \\ 
        \cmidrule{2-11}
        & \multirow{2}{*}{ Overall } & \multicolumn{2}{|c|}{ Question Type } & \multicolumn{2}{c|}{ Answer Type } & \multirow{2}{*}{ Overall } & \multicolumn{2}{|c|}{ Question Type } & \multicolumn{2}{c}{ Answer Type } \\
        \cmidrule{3-6}
        \cmidrule{8-11}
       & &\multicolumn{1}{|c|}{Complex} & Simple & \multicolumn{1}{|c|}{Entity} & \multicolumn{1}{c|}{Time} & & \multicolumn{1}{|c|}{Complex} & Simple & \multicolumn{1}{|c|}{Entity} & 
       \multicolumn{1}{c}{Time} \\
        \midrule
        BERT & 0.243 & 0.239 & 0.249 & 0.277 & 0.179 & 0.620 & 0.598 & 0.649 & 0.628 & 0.604 \\
        \midrule
        EmbedKGQA & 0.288 & 0.286 & 0.290 & 0.411 & 0.057 & 0.672 & 0.632 & 0.725 & 0.850 & 0.341 \\
        CronKGQA & 0.647 & 0.392 & 0.987 & 0.699 & 0.549 & 0.884 & 0.802 & 0.992 & 0.898 & 0.857 \\
        TMA & 0.784 & 0.632 & 0.987 &0.792 &0.743&0.943 & 0.904 &\textbf{0.995} &0.947 & 0.936\\
        TempoQR & \underline{0.918} & \underline{0.864} & \textbf{0.990} & \textbf{0.926} & \underline{0.903} & \textbf{0.978} & \textbf{0.967} & \underline{0.993} & \textbf{0.980} & \underline{0.974} \\
        \midrule
        Prog-TQA & \textbf{0.937} & \textbf{0.898} & \underline{0.989} & \underline{0.914} & \textbf{0.982} & \underline{0.973} & \underline{0.960} & \underline{0.993} & \underline{0.968} & \textbf{0.982} \\
        \bottomrule
        \end{tabular}
        }
    \caption{Comparison of various baselines and Prog-TQA on CronQuestions.
    }
    \vspace{-1em}
    \label{table:cron}
\end{table*}
\section{Experiments}
\subsection{Experimental Setup}
In this section, we first introduce the datasets used for evaluation. Then, we present baseline methods for comparison and finally explain the implementation details.

\textbf{Datasets.} 
We conduct experiments on two widely-used datasets for the TKGQA task: MultiTQ~\cite{chen2023multi} and CronQuestions~\cite{saxena2021question}. In MultiTQ, the time information is represented as time points, and temporal questions may be of single-constraint and multiple-constraints. In CronQuestions, the time information is represented as time intervals, and the questions are single-constraint only. The detailed statistics of these datasets can be found in Appendix ~\ref{app:dataset}.

\textbf{Baselines.}
We compare Prog-TQA with three types of baselines: 1) The embedding-based methods, including EmbedKGQA~\cite{saxena2020improving} and CronKGQA\cite{saxena2021question}. 2) The temporal-enhanced methods, including TempoQR\cite{mavromatis2022tempoqr}, MultiQA\cite{chen2023multi}, and TMA \cite{liu2023time}. 3) Additionally, the LM-based methods including BERT\cite{kenton2019bert}. The detailed analysis and comparison of these baselines can be found in Appendix \ref{app:baseline}.

\textbf{Implementation Details.}
In the draft generation module, we use vicuna-13B~\cite{vicuna2023} as the base LLM. We employ 8 examples for CronQuestions and 6 examples for MultiTQ, generating 1 program draft for each question. In the linking module, we directly utilize the entity and relation annotations given in the CronQuestions. For MultiTQ, we employ the miniLM~\cite{wang2020minilm} as the sentence encoder and retrieved 5 facts, linking the top 1 entity and 2 relations for each question. In the execution module, we activate the post-process submodule for the CronQuestions dataset but not for the MultiTQ dataset. In the self-improvement stage, we apply the Low-Rank Adaptation (LoRA)~\cite{hu2021lora} method for fine-tuning, with the rank set to 8. Experiments are conducted on 2 NVIDIA RTX3090 GPUs. Results are reported for 2 rounds of iteration on 100k fine-tuning data for MultiTQ and 1 round of iteration on 50k fine-tuning data for CronQuestions.
\subsection{Main Results}
Table~\ref{table:multiQA} and Table~\ref{table:cron} present the experimental results of Prog-TQA in comparison with other methods on the MultiTQ and CronQuestions datasets, where the highest results are highlighted in bold font and the second highest results are marked underlined. We report the Hits@1 and Hits@10 of the evaluation results.

As shown in Table~\ref{table:multiQA}, Prog-TQA achieves state-of-the-art performance across all question types for both hits@1 and hits@10 on MultiTQ. Specifically, compared to the previous state-of-the-art model, the hits@1 performances achieve 60.9\% improvement on multiple-constraint questions and 47.0\% improvement on single-constraint questions. The results demonstrate the capability of Prog-TQA for precisely answering temporal questions, especially for complex questions. 

As for CronQuestions, 
Prog-TQA still achieves a 2.1\% relative improvement on Hits@1 and competitive results on Hits@10. Given that CronQuestions contains only single-constraint questions that have relatively simple semantics, existing models have already achieved good performance on it. To this end, the improvement gains from Prog-TQA might not appear obvious. We believe that Prog-TQA would exhibit greater potential when faced with questions with multiple time constraints.

\begin{figure}
    \centering
    \includegraphics[width=\linewidth]{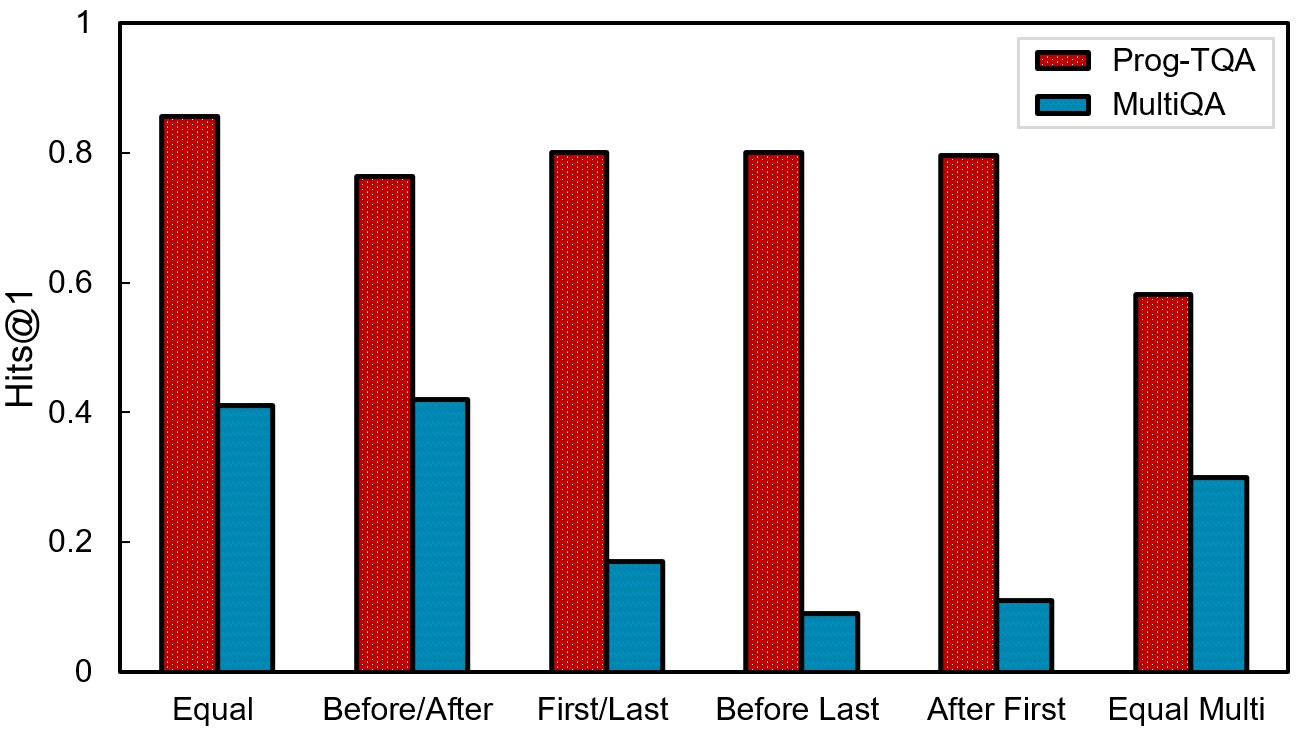}
    \caption{Performances (Hits@1) of Prog-TQA and MultiQA on the MultiTQ dataset against question types.}
    \label{fig:question-type}
    \vspace{-1em}
\end{figure}

Besides, we also conduct a detailed analysis across different question types, as shown in Figure \ref{fig:question-type}.  
The simplest type of ``\textit{equal}'' questions are solved the best. Questions of the ``\textit{Before/After}'' and ``\textit{First/Last}'' types show similar performance, but the former lags slightly behind. This can be attributed to the fact that time information is implicitly present in entities for some ``\textit{before/after}'' type questions (e.g., the question mentioned in Section \ref{sec:intro}), which requires additional reasoning steps.
For multiple-constraint questions, ``\textit{Before Last}'' and ``\textit{After First}'' questions exhibit only slightly lower performance compared to ``\textit{First/Last}'' questions, indicating the reasoning ability of Prog-TQA. However, the performance for ``\textit{Equal Multi}'' type questions is less satisfactory. For these questions with unique answers, the selected $r$ relations in the linking module generate redundant answers, thereby reducing the performance on Hits@1.

\subsection{Ablation Study}
To elaborate on the effectiveness of each part of Prog-TQA, we first perform the full model with the self-improvement strategy (Prog-TQA w/ SI). Then, we consider four settings in the ablation experiments on the MultiTQ dataset:  we removed the self-improvement strategy (Prog-TQA). Based on the Prog-TQA, we subsequently replaced the linking module with the one used in MultiQA (Prog-TQA w/o L), enabled the post-processing module (Prog-TQA w/ PP), and reduced the number of demonstration examples to one (Prog-TQA w/ D(1)). The results are shown in Table \ref{table:ablation}.

\textbf{Impact of self-improvement strategy.} 
Comparing the results of Prog-TQA w/ SI with Prog-TQA, there is a significant drop in overall results, especially on multiple-constraints questions. It demonstrates that the self-improvement strategy bootstraps the capabilities of Prog-TQA in effectively parsing and reasoning over questions, especially questions with multiple time constraints.

\textbf{Impact of linking module.} When comparing the results of Prog-TQA and Prog-TQA w/o L, we can observe that by replacing the linking module, the overall performances decrease 13.2\% on hits@1 and 4.3\% on hits@10, which indicates that the proposed linking method achieves accurate entity and relation recognition, and subsequently ensures the precise execution of generated programs.

\textbf{Impact of post-processing.} Comparing the Prog-TQA and Prog-TQA w/ PP, enabling the post-processing resulted in a notable general decline in Hits@1 results, but an improvement in Hits@10 results. 
This revealed that the post-processing process can greatly expand the recall rate of candidate answers, which increases performances by generating multiple programs for a draft, but at the cost of performance degradation on the more accurate Hits@1 metric. Therefore, we use the post-processing operation in the self-improvement to produce more programs for filtering high-quality annotations.

\textbf{Impact of the number of examples.} Finally, comparing the Prog-TQA with Prog-TQA w/ D(1), we observe that reducing the number of demonstrations leads to a significant decline in results. Although utilizing only one demonstration can provide a basic understanding of the KoPL language format to the LLM to boost the performances, it's crucial to employ more high-quality demonstrations to enhance LLM's abilities.
\begin{table}
    \resizebox{\linewidth}{!}{
    \begin{tabular}{l | c c c c c c }
        \toprule
        \multirow{2}{*}{\textbf{Model}} & \multicolumn{3}{c|}{ Hits@1 } & \multicolumn{3}{c}{ Hits@10 } \\
        \cmidrule{2-7}
        &Overall &\multicolumn{1}{|c|}{Multiple} & \multicolumn{1}{c|}{Single} & overall& \multicolumn{1}{|c|}{Multiple} & Single \\
        \midrule
        Prog-TQA w/ SI & \textbf{0.797} & \textbf{0.750} & \textbf{0.817} & \textbf{0.934} & \textbf{0.910} & \textbf{0.944}\\
        Prog-TQA & 0.583 &0.379  &0.666 &0.697 &0.466 &0.790\\
        \midrule
        Prog-TQA w/o L & 0.451 &0.250  &0.532 &0.654 &0.522 &0.708 \\
        Prog-TQA w/ PP & 0.537 &0.360  &0.609 &0.777 &0.682 &0.815 \\
        Prog-TQA w/ D(1) & 0.330 &0.252  &0.361 &0.498 &0.488 &0.502 \\
        \bottomrule
    \end{tabular}
        }
        \vspace{-0.5em}
     \caption{Results of the ablation study. “w/” means removing the module and “w/o” means adding the module.}
    \label{table:ablation}
\end{table}
\begin{table}
    \centering
    \resizebox{0.8\linewidth}{!}{
    \begin{tabular}{c  c }
        \toprule
        Round & Spurious program rate\\
          \midrule
          1st Iteration & 7.5\% \\
          \midrule
          2nd Iteration & 4.0\% \\
        \bottomrule
    \end{tabular}
        }
     \caption{Spurious program rate in the fine-tuning data at each round.}
     \vspace{-0.5em}
    \label{table:round}
\end{table}
\subsection{Further Analysis}
To conduct a comprehensive analysis of the key self-improvement strategy, we explore the impact of the number of fine-tuning rounds and the scale of fine-tuning data on the self-improvement process. Going further, we explore the distribution of the fine-tuning data with gold answers as weak supervision.

\begin{figure}
    \centering
    \includegraphics[width=1\linewidth]{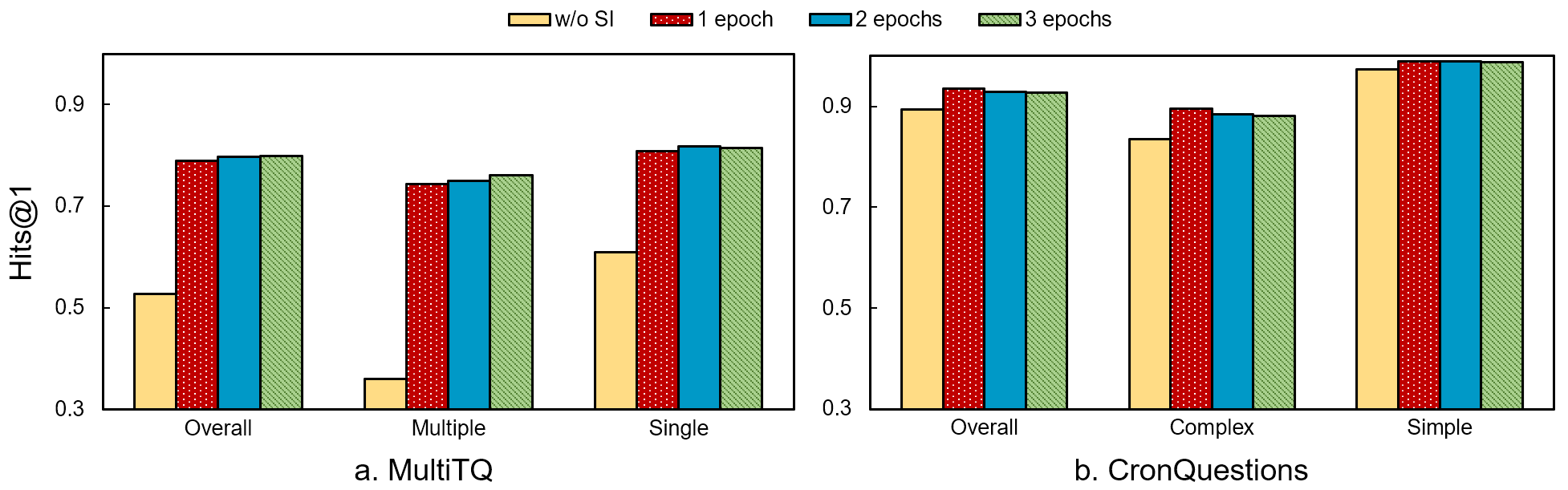}
    \caption{Performances (Hits@1) for each round of iterative fine-tuning on MultiTQ and CronQuestions. ``w/o SI'' indicates removing the self-improvement strategy.}
        \label{fig:fine-tune-epoch}
\end{figure}

\textbf{Number of iterative fine-tuning rounds.}
We analyze the performances on each round of iterative fine-tuning on two datasets, the results are shown in Figure \ref{fig:fine-tune-epoch}. 
Since Prog-TQA gradually learns from programs of relatively easy questions and conducts self-correction when handling challenging ones, more iterations are necessary to master complex questions.
We can observe that the best performance is achieved in the second round on MultiTQ and the first round on CronQuestions which contains only single-constraint questions.

It is worth noting that the performances no longer increase in the later rounds on CronQuestions. This is because the self-improvement procedure is affected by the distribution of simple and complex questions in the fine-tuning samples. 
As the iteration continues, the proportion of simple questions is too large for Prog-TQA to learn about the reasoning of complex ones.

\begin{figure}
    \centering
    \includegraphics[width=.95\linewidth]{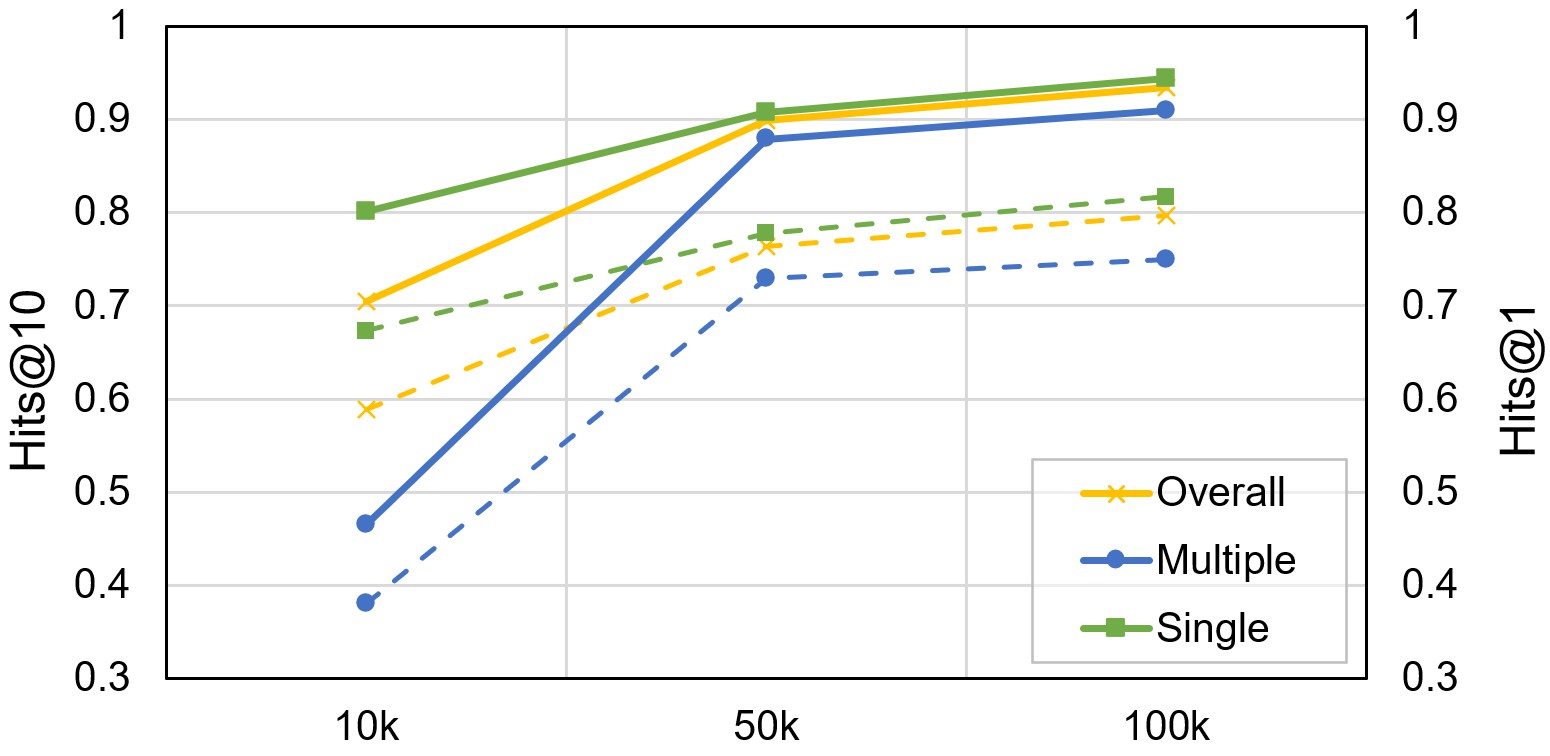}
    \caption{Performances under different sizes of fine-tuning data. The solid and dash lines donate Hits@10 and Hits@1 metrics, respectively.}
    \vspace{-0.5em}
    \label{fig:fine-tune}
\end{figure}
\textbf{Size of fine-tuning data.}
Figure \ref{fig:fine-tune} illustrates the results of iterative fine-tuning on 10k, 50k, and 100k of the training data. The performance of Prog-TQA steadily improves as the size of the dataset increases. Notably, the overall performances are relatively lower in the 10k dataset, because there are no sufficient correct annotations of multiple-type questions for Prog-TQA to learn from in a relatively small dataset. 

\textbf{Program distribution in fine-tuning data.} 
In the self-improvement strategy, we use correct answers as weak supervision to assess the generated programs. However, there may arise programs where the final answer is correct but the reasoning process is not, called spurious programs. Such programs may mislead the LLM and hinder it from learning complex reasoning steps. To understand the distribution of fine-tuning data, we analyze 200 randomly sampled programs from the first and second iterations of the MultiTQ dataset, respectively.  Given the absence of annotated programs, we manually evaluate them for spuriousness by assessing the correctness of entities, relations, and consistency of program logic with time constraints. Results in Table \ref{table:round} show a 7.5\% spurious program ratio for the first iteration and 4.0\% for the second.  This relatively low ratio attests to the effectiveness of our proposed self-improvement strategy, even considering the potential for iterative fine-tuning noise. 

\subsection{Case Study}
\begin{figure}
    \centering
    \includegraphics[width=\linewidth]{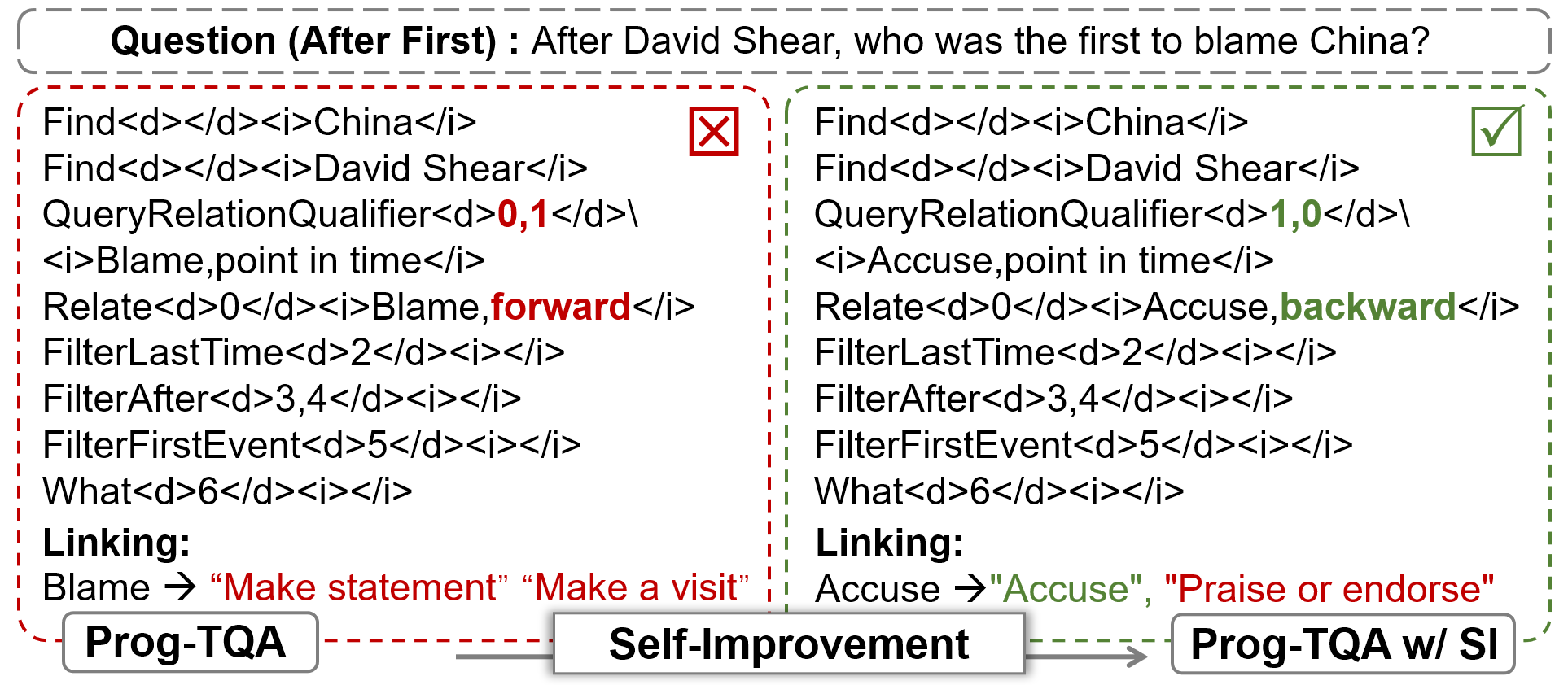}
    \caption{Drafts generated by Prog-TQA without the self-improvement strategy (Prog-TQA), and the complete Prog-TQA (Prog-TQA w/ SI).}
    \vspace{-0.5em}
    \label{fig:case}
\end{figure}

Figure~\ref{fig:case} illustrates the program drafts generated by Prog-TQA for the same question before and after the self-improvement strategy. It demonstrates the crucial role of self-improvement in helping the model understand temporal questions and generating accurate programs regarding time constraints. In the left draft, Prog-TQA fails to figure out the active and passive meaning of the relational parameters, resulting in two parameter assignment errors (red font). Besides, the LLM boosts its ability to link relations with self-generated high-quality programs in self-improvement. As shown in the lower part of Figure~\ref{fig:case}, Prog-TQA w/ SI can directly give a relation similar to the one in TKG, while Prog-TQA fails.
\section{Conclusion}
In this paper, we systematically analyzed the possible time constraints and designed corresponding temporal operators. Based on them, we proposed a semantic-parsing-based TKGQA framework called Prog-TQA. We further enhanced Prog-TQA's ability to understand temporal questions with an effective self-improvement strategy.
Prog-TQA first leverages the ICL ability of LLMs to generate KoPL program drafts. Subsequently, it utilizes the linking module to align the drafts with TKG and executes them with the execution module to generate answers. To further enhance the comprehension of the temporal questions, Prog-TQA incorporates an effective self-improvement strategy that iteratively bootstraps LLMs with high-quality self-generated annotations. Experimental results demonstrate that Prog-TQA can comprehensively understand the semantics of questions involving time constraints, and achieves significant performance gains on multiple-constraints questions.

\section{Acknowledgements}
This work is supported by NSFC No. 62372430, NSFC No. 62206266, and the Youth Innovation Promotion Association CAS No.2023112. We thank anonymous reviewers for their insightful comments and suggestions.

\nocite{*}
\section{Bibliographical References}\label{sec:reference}

\bibliographystyle{lrec-coling2024-natbib}
\bibliography{lrec-coling2024-example}

\begin{thebibliography}{36}
\expandafter\ifx\csname natexlab\endcsname\relax\def\natexlab#1{#1}\fi

\bibitem[{Brown et~al.(2020)Brown, Mann, Ryder, Subbiah, Kaplan, Dhariwal, Neelakantan, Shyam, Sastry, Askell et~al.}]{brown2020language}
Tom Brown, Benjamin Mann, Nick Ryder, Melanie Subbiah, Jared~D Kaplan, Prafulla Dhariwal, Arvind Neelakantan, Pranav Shyam, Girish Sastry, Amanda Askell, et~al. 2020.
\newblock Language models are few-shot learners.
\newblock \emph{Advances in neural information processing systems}, 33:1877--1901.

\bibitem[{Cao et~al.(2022)Cao, Shi, Pan, Nie, Xiang, Hou, Li, He, and Zhang}]{cao2022kqa}
Shulin Cao, Jiaxin Shi, Liangming Pan, Lunyiu Nie, Yutong Xiang, Lei Hou, Juanzi Li, Bin He, and Hanwang Zhang. 2022.
\newblock Kqa pro: A dataset with explicit compositional programs for complex question answering over knowledge base.
\newblock In \emph{Proceedings of the 60th Annual Meeting of the Association for Computational Linguistics (Volume 1: Long Papers)}, pages 6101--6119.

\bibitem[{Chen et~al.(2021)Chen, Liu, Yu, Lin, Lou, and Jiang}]{chen2021retrack}
Shuang Chen, Qian Liu, Zhiwei Yu, Chin-Yew Lin, Jian-Guang Lou, and Feng Jiang. 2021.
\newblock Retrack: A flexible and efficient framework for knowledge base question answering.
\newblock In \emph{Proceedings of the 59th annual meeting of the association for computational linguistics and the 11th international joint conference on natural language processing: system demonstrations}, pages 325--336.

\bibitem[{Chen et~al.(2023)Chen, Liao, and Zhao}]{chen2023multi}
Ziyang Chen, Jinzhi Liao, and Xiang Zhao. 2023.
\newblock Multi-granularity temporal question answering over knowledge graphs.
\newblock In \emph{Proceedings of the 61st Annual Meeting of the Association for Computational Linguistics (Volume 1: Long Papers)}, pages 11378--11392.

\bibitem[{Chiang et~al.(2023)Chiang, Li, Lin, Sheng, Wu, Zhang, Zheng, Zhuang, Zhuang, Gonzalez, Stoica, and Xing}]{vicuna2023}
Wei-Lin Chiang, Zhuohan Li, Zi~Lin, Ying Sheng, Zhanghao Wu, Hao Zhang, Lianmin Zheng, Siyuan Zhuang, Yonghao Zhuang, Joseph~E. Gonzalez, Ion Stoica, and Eric~P. Xing. 2023.
\newblock \href {https://lmsys.org/blog/2023-03-30-vicuna/} {Vicuna: An open-source chatbot impressing gpt-4 with 90\%* chatgpt quality}.

\bibitem[{Ding et~al.(2022)Ding, Chen, Li, and Qu}]{ding2022semantic}
Wentao Ding, Hao Chen, Huayu Li, and Yuzhong Qu. 2022.
\newblock Semantic framework based query generation for temporal question answering over knowledge graphs.
\newblock In \emph{Proceedings of the 2022 Conference on Empirical Methods in Natural Language Processing}, pages 1867--1877.

\bibitem[{Gu and Su(2022)}]{gu2022arcaneqa}
Yu~Gu and Yu~Su. 2022.
\newblock Arcaneqa: Dynamic program induction and contextualized encoding for knowledge base question answering.
\newblock In \emph{Proceedings of the 29th International Conference on Computational Linguistics}, pages 1718--1731.

\bibitem[{Hu et~al.(2021)Hu, Shen, Wallis, Allen-Zhu, Li, Wang, Wang, and Chen}]{hu2021lora}
Edwarllorad~J Hu, Yelong Shen, Phillip Wallis, Zeyuan Allen-Zhu, Yuanzhi Li, Shean Wang, Lu~Wang, and Weizhu Chen. 2021.
\newblock Lora: Low-rank adaptation of large language models.
\newblock \emph{arXiv preprint arXiv:2106.09685}.

\bibitem[{Huang et~al.(2023)Huang, Gu, Hou, Wu, Wang, Yu, and Han}]{huang2023large}
Jiaxin Huang, Shixiang Gu, Le~Hou, Yuexin Wu, Xuezhi Wang, Hongkun Yu, and Jiawei Han. 2023.
\newblock Large language models can self-improve.
\newblock In \emph{Proceedings of the 2023 Conference on Empirical Methods in Natural Language Processing}, pages 1051--1068.

\bibitem[{Izacard et~al.(2022)Izacard, Lewis, Lomeli, Hosseini, Petroni, Schick, Dwivedi-Yu, Joulin, Riedel, and Grave}]{izacard2022few}
Gautier Izacard, Patrick Lewis, Maria Lomeli, Lucas Hosseini, Fabio Petroni, Timo Schick, Jane Dwivedi-Yu, Armand Joulin, Sebastian Riedel, and Edouard Grave. 2022.
\newblock Few-shot learning with retrieval augmented language models.
\newblock \emph{arXiv preprint arXiv:2208.03299}.

\bibitem[{Izacard et~al.(2023)Izacard, Lewis, Lomeli, Hosseini, Petroni, Schick, Dwivedi-Yu, Joulin, Riedel, and Grave}]{baek2023knowledge}
Gautier Izacard, Patrick Lewis, Maria Lomeli, Lucas Hosseini, Fabio Petroni, Timo Schick, Jane Dwivedi-Yu, Armand Joulin, Sebastian Riedel, and Edouard Grave. 2023.
\newblock Atlas: Few-shot learning with retrieval augmented language models.
\newblock \emph{Journal of Machine Learning Research}, 24(251):1--43.

\bibitem[{Jia et~al.(2018{\natexlab{a}})Jia, Abujabal, Saha~Roy, Str{\"o}tgen, and Weikum}]{jia2018tempquestions}
Zhen Jia, Abdalghani Abujabal, Rishiraj Saha~Roy, Jannik Str{\"o}tgen, and Gerhard Weikum. 2018{\natexlab{a}}.
\newblock Tempquestions: A benchmark for temporal question answering.
\newblock In \emph{Companion Proceedings of the The Web Conference 2018}, pages 1057--1062.

\bibitem[{Jia et~al.(2018{\natexlab{b}})Jia, Abujabal, Saha~Roy, Str{\"o}tgen, and Weikum}]{jia2018tequila}
Zhen Jia, Abdalghani Abujabal, Rishiraj Saha~Roy, Jannik Str{\"o}tgen, and Gerhard Weikum. 2018{\natexlab{b}}.
\newblock Tequila: Temporal question answering over knowledge bases.
\newblock In \emph{Proceedings of the 27th ACM international conference on information and knowledge management}, pages 1807--1810.

\bibitem[{Jia et~al.(2021)Jia, Pramanik, Saha~Roy, and Weikum}]{jia2021complex}
Zhen Jia, Soumajit Pramanik, Rishiraj Saha~Roy, and Gerhard Weikum. 2021.
\newblock Complex temporal question answering on knowledge graphs.
\newblock In \emph{Proceedings of the 30th ACM international conference on information \& knowledge management}, pages 792--802.

\bibitem[{Kenton and Toutanova(2019)}]{kenton2019bert}
Jacob Devlin Ming-Wei~Chang Kenton and Lee~Kristina Toutanova. 2019.
\newblock Bert: Pre-training of deep bidirectional transformers for language understanding.
\newblock In \emph{Proceedings of NAACL-HLT}, pages 4171--4186.

\bibitem[{Lacroix et~al.(2019)Lacroix, Obozinski, and Usunier}]{lacroix2019tensor}
Timoth{\'e}e Lacroix, Guillaume Obozinski, and Nicolas Usunier. 2019.
\newblock Tensor decompositions for temporal knowledge base completion.
\newblock In \emph{International Conference on Learning Representations}.

\bibitem[{Lan et~al.(2021)Lan, He, Jiang, Jiang, Zhao, and Wen}]{LanHJ0ZW21}
Yunshi Lan, Gaole He, Jinhao Jiang, Jing Jiang, Wayne~Xin Zhao, and Ji{-}Rong Wen. 2021.
\newblock \href {https://doi.org/10.24963/ijcai.2021/611} {A survey on complex knowledge base question answering: Methods, challenges and solutions}.
\newblock In \emph{Proceedings of the Thirtieth International Joint Conference on Artificial Intelligence, {IJCAI} 2021, Virtual Event / Montreal, Canada, 19-27 August 2021}, pages 4483--4491. ijcai.org.

\bibitem[{Li et~al.(2023)Li, Ma, Zhuang, Gu, Su, and Chen}]{li2023few}
Tianle Li, Xueguang Ma, Alex Zhuang, Yu~Gu, Yu~Su, and Wenhu Chen. 2023.
\newblock Few-shot in-context learning on knowledge base question answering.
\newblock In \emph{Proceedings of the 61st Annual Meeting of the Association for Computational Linguistics (Volume 1: Long Papers)}, pages 6966--6980.

\bibitem[{Liu et~al.(2023)Liu, Liang, Fang, Wang, Wu, and Jiang}]{liu2023time}
Yonghao Liu, Di~Liang, Fang Fang, Sirui Wang, Wei Wu, and Rui Jiang. 2023.
\newblock Time-aware multiway adaptive fusion network for temporal knowledge graph question answering.
\newblock In \emph{ICASSP 2023-2023 IEEE International Conference on Acoustics, Speech and Signal Processing (ICASSP)}, pages 1--5. IEEE.

\bibitem[{L{\"u}demann et~al.(2020)L{\"u}demann, Shiba, Thymianis, Heist, Ludwig, and Paulheim}]{ludemann2020knowledge}
Niklas L{\"u}demann, Ageda Shiba, Nikolaos Thymianis, Nicolas Heist, Christopher Ludwig, and Heiko Paulheim. 2020.
\newblock A knowledge graph for assessing agressive tax planning strategies.
\newblock In \emph{The Semantic Web--ISWC 2020: 19th International Semantic Web Conference, Athens, Greece, November 2--6, 2020, Proceedings, Part II 19}, pages 395--410. Springer.

\bibitem[{Mavromatis et~al.(2022)Mavromatis, Subramanyam, Ioannidis, Adeshina, Howard, Grinberg, Hakim, and Karypis}]{mavromatis2022tempoqr}
Costas Mavromatis, Prasanna~Lakkur Subramanyam, Vassilis~N Ioannidis, Adesoji Adeshina, Phillip~R Howard, Tetiana Grinberg, Nagib Hakim, and George Karypis. 2022.
\newblock Tempoqr: Temporal question reasoning over knowledge graphs.
\newblock In \emph{36th AAAI Conference on Artificial Intelligence, AAAI 2022}, pages 5825--5833. Association for the Advancement of Artificial Intelligence.

\bibitem[{Meng et~al.(2023)Meng, Michalski, Huang, Zhang, Abdelzaher, and Han}]{meng2023tuning}
Yu~Meng, Martin Michalski, Jiaxin Huang, Yu~Zhang, Tarek Abdelzaher, and Jiawei Han. 2023.
\newblock Tuning language models as training data generators for augmentation-enhanced few-shot learning.
\newblock In \emph{International Conference on Machine Learning}, pages 24457--24477. PMLR.

\bibitem[{Nie et~al.(2023)Nie, Zhang, Wang, and Liu}]{nie2023code}
Zhijie Nie, Richong Zhang, Zhongyuan Wang, and Xudong Liu. 2023.
\newblock Code-style in-context learning for knowledge-based question answering.
\newblock \emph{arXiv preprint arXiv:2309.04695}.

\bibitem[{Qian et~al.(2023)Qian, Chen, Cong, Xu, and Wang}]{qian}
Tangwen Qian, Yile Chen, Gao Cong, Yongjun Xu, and Fei Wang. 2023.
\newblock \href {https://doi.org/10.48550/ARXIV.2312.14394} {Adaptraj: {A} multi-source domain generalization framework for multi-agent trajectory prediction}.
\newblock \emph{CoRR}, abs/2312.14394.

\bibitem[{Saxena et~al.(2021)Saxena, Chakrabarti, and Talukdar}]{saxena2021question}
A~Saxena, S~Chakrabarti, and P~Talukdar. 2021.
\newblock Question answering over temporal knowledge graphs.
\newblock In \emph{ACL-IJCNLP 2021-59th Annual Meeting of the Association for Computational Linguistics and the 11th International Joint Conference on Natural Language Processing, Proceedings of the Conference}, pages 6663--6676. Association for Computational Linguistics (ACL).

\bibitem[{Saxena et~al.(2020)Saxena, Tripathi, and Talukdar}]{saxena2020improving}
Apoorv Saxena, Aditay Tripathi, and Partha Talukdar. 2020.
\newblock Improving multi-hop question answering over knowledge graphs using knowledge base embeddings.
\newblock In \emph{Proceedings of the 58th annual meeting of the association for computational linguistics}, pages 4498--4507.

\bibitem[{Shang et~al.(2022)Shang, Wang, Qi, and Huang}]{shang2022improving}
Chao Shang, Guangtao Wang, Peng Qi, and Jing Huang. 2022.
\newblock Improving time sensitivity for question answering over temporal knowledge graphs.
\newblock In \emph{Proceedings of the 60th Annual Meeting of the Association for Computational Linguistics (Volume 1: Long Papers)}, pages 8017--8026.

\bibitem[{Shao et~al.(2022)Shao, Zhang, Wang, and Xu}]{shao}
Zezhi Shao, Zhao Zhang, Fei Wang, and Yongjun Xu. 2022.
\newblock \href {https://doi.org/10.1145/3534678.3539396} {Pre-training enhanced spatial-temporal graph neural network for multivariate time series forecasting}.
\newblock In \emph{{KDD} '22: The 28th {ACM} {SIGKDD} Conference on Knowledge Discovery and Data Mining, Washington, DC, USA, August 14 - 18, 2022}, pages 1567--1577. {ACM}.

\bibitem[{Shu et~al.(2022)Shu, Yu, Li, Karlsson, Ma, Qu, and Lin}]{shu2022tiara}
Yiheng Shu, Zhiwei Yu, Yuhan Li, B{\"o}rje Karlsson, Tingting Ma, Yuzhong Qu, and Chin-Yew Lin. 2022.
\newblock Tiara: Multi-grained retrieval for robust question answering over large knowledge base.
\newblock In \emph{Proceedings of the 2022 Conference on Empirical Methods in Natural Language Processing}, pages 8108--8121.

\bibitem[{Tan et~al.(2023)Tan, Chen, Shao, and Chen}]{tan2023make}
Chuanyuan Tan, Yuehe Chen, Wenbiao Shao, and Wenliang Chen. 2023.
\newblock Make a choice! knowledge base question answering with in-context learning.
\newblock \emph{arXiv preprint arXiv:2305.13972}.

\bibitem[{Touvron et~al.(2023)Touvron, Martin, Stone, Albert, Almahairi, Babaei, Bashlykov, Batra, Bhargava, Bhosale et~al.}]{touvron2023llama}
Hugo Touvron, Louis Martin, Kevin Stone, Peter Albert, Amjad Almahairi, Yasmine Babaei, Nikolay Bashlykov, Soumya Batra, Prajjwal Bhargava, Shruti Bhosale, et~al. 2023.
\newblock Llama 2: Open foundation and fine-tuned chat models.
\newblock \emph{arXiv preprint arXiv:2307.09288}.

\bibitem[{Wang et~al.(2023)Wang, Yao, Li, Sun, and Zhang}]{wang2023ai}
Fei Wang, Di~Yao, Yong Li, Tao Sun, and Zhao Zhang. 2023.
\newblock Ai-enhanced spatial-temporal data-mining technology: New chance for next-generation urban computing.
\newblock \emph{The Innovation}, 4(2).

\bibitem[{Wang et~al.(2020)Wang, Wei, Dong, Bao, Yang, and Zhou}]{wang2020minilm}
Wenhui Wang, Furu Wei, Li~Dong, Hangbo Bao, Nan Yang, and Ming Zhou. 2020.
\newblock Minilm: Deep self-attention distillation for task-agnostic compression of pre-trained transformers.
\newblock \emph{Advances in Neural Information Processing Systems}, 33:5776--5788.

\bibitem[{Xu et~al.(2023)Xu, Wang, An, Wang, and Zhang}]{xu2023artificial}
Yongjun Xu, Fei Wang, Zhulin An, Qi~Wang, and Zhao Zhang. 2023.
\newblock Artificial intelligence for science—bridging data to wisdom.
\newblock \emph{The Innovation}, 4(6).

\bibitem[{Yu et~al.(2021)Yu, Zuo, Jiang, Ren, Zhao, and Zhang}]{yu2021fine}
Yue Yu, Simiao Zuo, Haoming Jiang, Wendi Ren, Tuo Zhao, and Chao Zhang. 2021.
\newblock Fine-tuning pre-trained language model with weak supervision: A contrastive-regularized self-training approach.
\newblock In \emph{Proceedings of the 2021 Conference of the North American Chapter of the Association for Computational Linguistics: Human Language Technologies}, pages 1063--1077.

\bibitem[{Zhang et~al.(2023)Zhang, Guan, Zhang, Zhuang, An, Wang, and Xu}]{zhang2023weighted}
Zhao Zhang, Zhanpeng Guan, Fuwei Zhang, Fuzhen Zhuang, Zhulin An, Fei Wang, and Yongjun Xu. 2023.
\newblock Weighted knowledge graph embedding.
\newblock In \emph{Proceedings of the 46th International ACM SIGIR Conference on Research and Development in Information Retrieval}, pages 867--877.

\end{thebibliography}

\appendix
\section{Appendix}
\subsection{Details about KoPL and Temporal Operators}\label{app:tempop}

In this section, we present details regarding the KoPL programming language, the simplified KoPL format, and the design of temporal operators.

\textbf{KoPL}: KoPL\cite{cao2022kqa} is a compositional and interpretable programming language, models the complex procedure of question answering through a program comprising intermediate steps. Each step consists a function containing a list of textual arguments $a_i$ and a list of function arguments $b_i$. The fucntion arguments represent the function list, indicating from which previous functions the current function receives its input.
For example, the function ``\textit{Relate(Find(Sudan), Make a visit, forward)}'' has two textual inputs: relation ``\textit{Make a visit}'' and direction ``\textit{forward}'' and one functional input ``\textit{Find(Sudan)}'' derived from the preceding function ``\textit{Find}''.

A KoPL program consists of a sequence of KoPL functions, presenting a tree-like program structure, which can be serialized into a sequence of functions by post-order traversal and executed.

\textbf{Simplified KoPL format}: In the draft generation module, we simplified the tree-like program representation. We keep the list of textual arguments and wrap them with <d></d>. For function arguments, we utilize the index representation of their order in the program, enclosing them with <i></i> tags. This format can be restored to the original program structure and executed in the same manner.

\textbf{Details about design of temporal operators}: 
The designed temporal operators are rooted in the KoPL function library and are consistent with KoPL function naming conventions while expressing the action semantics. In the design of three types of operators, aiming at TKGs with different time granularity and storage formats of time information, the fundamental temporal operators are compatible with all granularity and storage formats. 

To help the LLM understand the semantics of the question more clearly, in the design of the precise temporal operators, the temporal constraint of ``simultaneous" is deconstructed into three specific precise operations. These operations include retrieving facts utilizing coarse-grained time (FilterRange), time points (FilterByTimePoint), and time intervals (FilterByDuration). Similarly, the implementation of the FilterFirst (FilterLast) operator is segmented into FilterFirstTime (FilterLastTime) and FilterFirstEvent (FilterLastEvent) to help the LLM distinguish whether the query object of the function is a list of times or a list of facts.

\subsection{Baseline Methods}\label{app:baseline}
In the experiments, we compared Prog-TQA with the embedding-based, temporal-enhanced, and LM-based methods.

\textbf{The embedding-based methods} aim to learn the embedding of questions and candidate answers and predict the answers by embedding calculation. The methods try to introduce pre-train KG and TKG embedding models \cite{zhang2023weighted, lacroix2019tensor} to enhance the embeddings. EmbedKGQA~\cite{saxena2020improving} derives entity embeddings from pre-trained KG models, while CronKGQA~\cite{saxena2021question} extracts entity and time embeddings from TKG models and incorporates time-aware question embeddings. Both methods predict answers by computing matching scores between encoded questions and candidate answers.

\textbf{The temporal-enhanced methods} focus on identifying the time information and using it to enhance embeddings or retrieve relevant facts to address complex questions. TempoQR\cite{mavromatis2022tempoqr} fuses time information by retrieving time scopes of entities mentioned in the question. TMA\cite{liu2023time} enhances performance through improved fact retrieval and an adaptive fusion network.  MultiQA\cite{chen2023multi} extracts time texts from questions and enhances the time-aware embedding of questions with a multi-granularity representation module.

\textbf{The LM-based methods}. Additionally, we also utilized BERT\cite{kenton2019bert} to obtain embeddings of entities, time, and questions, which were concatenated to calculate the answer, allowing us to analyze the effect of LMs in handling temporal questions.

\subsection{Further analysis on the LLM}\label{app:llm}

\begin{table}[ht]
    \centering
    \resizebox{\linewidth}{!}{
        \begin{tabular}{l | c c c c c c }
        \toprule
        \multirow{2}{*}{\textbf{Model}} & \multicolumn{3}{c|}{ Hits@1 } & \multicolumn{3}{c}{ Hits@10 } \\ 
        \cmidrule{2-7}
        &Overall &\multicolumn{1}{|c|}{Multiple} & \multicolumn{1}{c|}{Single} & overall& \multicolumn{1}{|c|}{Multiple} & Single \\
        \midrule
        Llama-13B & 0.797 & 0.750 & 0.817 & 0.934 & 0.910 & 0.944  \\
        Vicuna-13B & 0.793 & 0.748 & 0.811 & 0.930 & 0.912 & 0.937  \\
        \bottomrule
        \end{tabular}
        }
 \caption{
 Performance of Prog-TQA with different LLM variants. }
    \label{table:variant}
\end{table}
\begin{table}[ht]
    \centering
    \resizebox{\linewidth}{!}{
        \begin{tabular}{l | c c c c c c }
        \toprule
        \multirow{2}{*}{\textbf{Model}} & \multicolumn{3}{c|}{ Hits@1 } & \multicolumn{3}{c}{ Hits@10 } \\ 
        \cmidrule{2-7}
        &Overall &\multicolumn{1}{|c|}{Multiple} & \multicolumn{1}{c|}{Single} & overall& \multicolumn{1}{|c|}{Multiple} & Single \\
        \midrule
        7B & 0.516 & 0.290 & 0.602 & 0.754 & 0.584 & 0.820  \\
        13B & 0.529 & 0.346 & 0.598 & 0.772 & 0.683 & 0.806  \\
        33B & 0.530 & 0.331 & 0.607 & 0.793 & 0.666 & 0.842  \\
        \bottomrule
        \end{tabular}
        }
 \caption{
 Performance of Prog-TQA with different scales of LLM. Due to cost constraints, experiments were conducted on randomly selected 5 subsets of size 2000 from MultiTQ, with average results being reported.}
    \label{table:vicuna-size}
\end{table}

To assess the effectiveness of the LLM in Prog-TQA, we analyze the impact of different LLM variants and LLM with different parameter scales on the performance of Prog-TQA on the MultiTQ dataset.

\textbf{Variants of LLM.} 
we conduct experiments using Llama-13B~\cite{touvron2023llama} as the base model on the MultiTQ dataset. The result can be found in Table \ref{table:variant}. It can be observed that the performances of the two models are very close. Notably, both models outperform all baselines on the MultiTQ dataset. The observation indicates that using different LLM variants with similar parameter scales as the base model in Prog-TQA doesn’t bring significant performance differences. 

\textbf{Scale of LLM.} 
We choose three LLMs which differ only in parameter scale to explore the influence of parameter scale on the performance of Prog-TQA. Due to the cost limitation of fine-tuning the 33B model, we use Prog-TQA without the self-improvement module to conduct experiments on the MultiTQ dataset. We activate the post-processing submodule to make full use of the generated programs.
The results are presented in Table \ref{table:vicuna-size}. The performance improves slightly when the scales grow exponentially. 
It shows that the parameter scale of LLMs is not a decisive factor in improving performance. Combined with the results of ablation experiments, we can deduce that the proposed linking module and post-processing in the execution module, as well as the critical self-improvement strategy, ensure the performance of Prog-TQA jointly.
\subsection{Dataset Statistics}\label{app:dataset}
We conduct experiments on the MultiTQ and CronQuestions datasets. The statistics for the two datasets are shown in Table~\ref{table:dataset}.
\begin{table*}[ht]
\resizebox{\linewidth}{!}{
\begin{tabular}{l | ccc | ccc |c|c|ccc|c}
\toprule
 & \multicolumn{7}{c|}{\textbf{MultiTQ}} & \multicolumn{5}{c}{\textbf{CronQuestions}} \\
\cmidrule{2-13}
 & \multicolumn{3}{c|}{Single} & \multicolumn{3}{c|}{Multiple} & \multirow{2}{*}{\textbf{Total}} & \multicolumn{1}{c|}{Simple} & \multicolumn{3}{c|}{Complex} & \multirow{2}{*}{\textbf{Total}} \\
\cmidrule{2-7} \cmidrule{9-12}
 & \multicolumn{1}{c}{Equal} & \multicolumn{1}{c}{Before/After} & \multicolumn{1}{c|}{First/Last} & \multicolumn{1}{c}{Equal Multi} & \multicolumn{1}{c}{After First} & \multicolumn{1}{c|}{Before Last} &  & \multicolumn{1}{c|}{Simple} & \multicolumn{1}{c}{Before/After} & \multicolumn{1}{c}{First/Last} & \multicolumn{1}{c|}{Time Join} &  \\
\midrule
\textbf{Train} & 135,890 & 75,340 & 72,252 & 16,893 & 43,305 & 43,107 & 386,787 & 152,122 & 23,869 & 118,556 & 55,453 & 350,000 \\
\textbf{Dev.} & 18,983 & 11,655 & 11,097 & 3,213 & 6,499 & 6,532 & 57,979 & 12,942 & 1,982 & 11,198 & 3,878 & 30,000 \\
\textbf{Test} & 17,311 & 11,073 & 10,480 & 3,207 & 6,266 & 6,247 & 54,584 & 12,858 & 2,151 & 11,159 & 3,832 & 30,000\\
\bottomrule
\end{tabular}
}
\caption{Statistic of question types on MultiTQ and CronQuestions datasets.}
\vspace{-1em}
\label{table:dataset}
\end{table*}
\subsection{Annotated Programs and Prompt Template}\label{app:annot}

\textbf{Prompt Template}. Figure \ref{table:draft} shows a prompt example in the draft generation module, which is used in the initial stage of the self-improvement strategy. Figure \ref{table:template} shows the prompt template used in the fine-tuning stage of the self-improvement strategy.

\textbf{Annotated Programs}. We present one representative question and corresponding manual annotations from each predefined question category in both datasets, as shown in Table \ref{table:annotation} and Table \ref{table:cron_annotation}.

\subsection{Difference with spatiotemporal graph reasoning task}
The spatiotemporal graph reasoning task aims at discovering useful patterns from the dynamic interactions of spatial and temporal information in the data. It plays an important role in various tasks related to mining spatio-temporal patterns such as time series~\cite{shao}, trajectories~\cite{qian}, and TKG. Notably, while both spatiotemporal graph reasoning and temporal question answering are crucial in TKG, they cater to unique challenges: TKGQA focuses on answering natural language questions by diving into the time constraints and retrieving relevant facts, whereas spatiotemporal reasoning intends to mine patterns of temporal changes in structured information to achieve effective querying of the information we are interested in.
\begin{figure*}
    \begin{mybox}{Prompt in Draft Generation Module}
\texttt{\#\#\# Instruction: Convert a natural language question to a KoPL query.} \\ \\
        \texttt{\#\#\# Input: In the same month, who appealed to Tony Blair on behalf of an Indian Member of Parliament?} \\
        \texttt{\#\#\# Response: Find<d></d><i>Tony Blair</i>}\\
        \texttt{Find<d></d><i>Member of Parliament (India)</i>}\\
        \texttt{QueryRelationQualifier<d>1,0</d><i>Make an appeal or request,point in time</i>}\\
        \texttt{Relate<d>0</d><i>Make an appeal or request,backward</i>}\\
        \texttt{FilterFirstTime<d>2</d><i></i>}\\
        \texttt{GetMonth<d>4</d><i></i>}\\
        \texttt{FilterRange<d>3,5</d><i></i>}\\
        \texttt{What<d>6</d><i></i>}\\ \\
        ...\\ \\
        \texttt{\#\#\# Input: With whom did Daniel Ortega want to negotiate in the same month as Dmitry Anatolyevich Medvedev?} \\
        \texttt{\#\#\# Response: Find<d></d><i>Daniel Ortega</i>}\\
        \texttt{Find<d></d><i>Dmitry Anatolyevich Medvedev</i>}\\
        \texttt{QueryRelationQualifier<d>0,1</d><i>Express intent to meet or negotiate,point in time</i>}\\
        \texttt{Relate<d>0</d><i>Express intent to meet or negotiate,forward</i>}\\
        \texttt{FilterFirstTime<d>2</d><i></i>}\\
        \texttt{GetMonth<d>4</d><i></i>}\\
        \texttt{FilterRange<d>3,5</d><i></i>}\\
        \texttt{What<d>6</d><i></i>}\\ \\
        \texttt{\#\#\# Input: Who visited China in the same month as Oleg Ostapenko?} \\
        \texttt{\#\#\# Response:} \\
\end{mybox}
\caption{The example prompt in the draft generation module, which is used in the initial stage of self-improvement.}
\label{table:draft}
\end{figure*}

\begin{figure*}
    \begin{mybox}{Template used in fine-tuning}
       \texttt{\#\#\# Instruction: Convert a natural language question to a KoPL query.} \\ \\
        \texttt{\#\#\# Input: \{\textbf{question}\}} \\
        \texttt{\#\#\# Response: \{\textbf{simplified KoPL program}\}}\\
\end{mybox}
\caption{The prompt template used in the fine-tuning stage of the self-improvement strategy.}
\label{table:template}
\end{figure*}
\begin{table*}[ht]
    \centering
    \resizebox{\textwidth}{!}{
        \begin{tabular}{ p{0.95\textwidth} }
        \toprule
         \textbf{\texttt{Equal}}: Who did Avelino J. Cruz, Jr. wish to negotiate with in the same month of Philippine military personnel? \\
        \midrule
        \texttt{Find<d></d><i>Avelino J. Cruz Jr.</i>}\\
        \texttt{Find<d></d><i>Military Personnel (Philippines)</i>}\\
        \texttt{QueryRelationQualifier<d>0,1</d><i>Express intent to meet or negotiate,point in time</i>}\\
        \texttt{Relate<d>0</d><i>Express intent to meet or negotiate,forward</i>}\\
        \texttt{FilterFirstTime<d>2</d><i></i>}\\
        \texttt{GetMonth<d>4</d><i></i>}\\
        \texttt{FilterRange<d>3,5</d><i></i>}\\
        \texttt{What<d>6</d><i></i>}\\
        \midrule
         \textbf{\texttt{Before/After}}: Before Opposition Supporter of Pakistan, who blamed  Militant of Taliban? \\
        \midrule
        \texttt{Find<d></d><i>Militant (Taliban)</i>}\\
        \texttt{Find<d></d><i>Opposition Supporter (Pakistan)</i>}\\
        \texttt{QueryRelationQualifier<d>1,0</d><i>Accuse,point in time</i>}\\
        \texttt{Relate<d>0</d><i>Accuse,backward</i>}\\
        \texttt{FilterFirstTime<d>2</d><i></i>}\\
        \texttt{FilterBefore<d>3,4</d><i></i>}\\
        \texttt{What<d>5</d><i></i>}\\
        \midrule
         \textbf{\texttt{First/Last}}: In which month did the Israeli police use conventional military force against the Israeli Defence Forces for the first time? \\
        \midrule
        \texttt{Find<d></d><i>Police (Israel)</i>}\\
        \texttt{Find<d></d><i>Israeli Defense Forces</i>}\\
        \texttt{QueryRelationQualifier<d>0,1</d><i>Use conventional military force,point in time</i>}\\
        \texttt{FilterFirstTime<d>2</d><i></i>}\\
        \texttt{GetMonth<d>3</d><i></i>}\\
        \midrule
         \textbf{\texttt{Equal-Multi}}: In 2014, against whom did the men of South Africa use unconventional violence for the first time? \\
        \midrule
        \texttt{Find<d></d><i>Men (South Africa)</i>}\\
        \texttt{Relate<d>0</d><i>Use unconventional violence,forward</i>}\\
        \texttt{FilterRange<d>1</d><i>2014</i>}\\
        \texttt{FilterFirstEvent<d>2</d><i></i>}\\
        \midrule
         \textbf{\texttt{Before Last}}: Before Macky Sall, who last negotiated with Barack Obama? \\
        \midrule
        \texttt{Find<d></d><i>Barack Obama</i>}\\
        \texttt{Find<d></d><i>Macky Sall</i>}\\
        \texttt{QueryRelationQualifier<d>1,0</d><i>Engage in negotiation,point in time</i>}\\
        \texttt{Relate<d>0</d><i>Engage in negotiation,backward</i>}\\
        \texttt{FilterFirstTime<d>2</d><i></i>}\\
        \texttt{FilterBefore<d>3,4</d><i></i>}\\
        \texttt{FilterLastEvent<d>5</d><i></i>}\\
        \midrule
         \textbf{\texttt{After First}}: After the Ministry of Information of Somalia, who was the first to visit Sudan? \\
        \midrule
        \texttt{Find<d></d><i>Sudan</i>}\\
        \texttt{Find<d></d><i>Information Ministry (Somalia)</i>}\\
        \texttt{QueryRelationQualifier<d>1,0</d><i>Make a visit,point in time</i>}\\
        \texttt{Relate<d>0</d><i>Make a visit,backward</i>}\\
        \texttt{FilterLastTime<d>2</d><i></i>}\\
        \texttt{FilterAfter<d>3,4</d><i></i>}\\
        \texttt{FilterFirstEvent<d>5</d><i></i>}\\
        \bottomrule
        \end{tabular}
        }
 \caption{The annotation examples for each question type in MultiTQ dataset.}
    \label{table:annotation}
\end{table*}

\begin{table*}[ht]
    \centering
    \resizebox{\textwidth}{!}{
        \begin{tabular}{ p{0.95\textwidth} }
        \toprule
         \textbf{\texttt{Simple}}: What team Angelo Buratti played in 1956? \\
        \midrule
        \texttt{Find<d></d><i>Angelo Buratti</i>}\\
        \texttt{Relate<d>0</d><i>member of sports team|forward</i>}\\
        \texttt{FilterByTimePoint<d>1</d><i>1956</i>}\\
        \texttt{What<d>2</d><i></i>}\\
        \midrule
         \textbf{\texttt{Before/After}}: Which person was sovereign of Navarre before 18th century \\
        \midrule
        \texttt{Find<d></d><i>monarch of Navarre</i>}\\
        \texttt{Relate<d>0</d><i>position held|backward</i>}\\
        \texttt{QueryEventQualifier<d></d><i>18th century|start time</i>}\\
        \texttt{FilterFirstTime<d>2</d><i></i>}\\
        \texttt{FilterBefore<d>1,3</d><i></i>}\\
        \texttt{What<d>4</d><i></i>}\\
        \midrule
         \textbf{\texttt{First/Last}}: When was John Salako playing their first play?\\
        \midrule
        \texttt{Find<d></d><i>John Salako</i>}\\
        \texttt{Relate<d>0</d><i>member of sports team|forward</i>}\\
        \texttt{FilterFirstTime<d>1</d><i></i>}\\
        \midrule
         \textbf{\texttt{Time-Join}}: Who was the Colorado Governor during Eurovision Song Contest 2005\\
        \midrule
        \texttt{Find<d></d><i>Governor of Colorado</i>}\\
        \texttt{Relate<d>0</d><i>position held|backward</i>}\\
        \texttt{QueryEventQualifier<d></d><i>Eurovision Song Contest 2005|duration</i>}\\
        \texttt{FilterByDuration<d>1,2</d><i></i>}\\
        \texttt{What<d>3</d><i></i>}\\
        \bottomrule
        \end{tabular}
        }
 \caption{The annotation examples for each question type in the CronQuestions dataset.}
    \label{table:cron_annotation}
\end{table*}
\end{document}